\documentclass{article}
\usepackage{spconf,amsmath,graphicx}
\usepackage{tabu}
\usepackage{amssymb}
\usepackage{subfig}
\usepackage{hyperref}

\usepackage{multirow}

\title{Puzzle-CAM: Improved localization via matching partial and full features}

\twoauthors
 {Sanghyun Jo}
	{GYNetworks\\
	\tt\small josanghyeokn@gynetworks.com}
 {In-Jae Yu\sthanks{Corresponding Author}}
	{KAIST \\ School of Computing \\
	\tt\small myhome98304@gmail.com}
	
\begin{document}
%
\maketitle
%

\begin{abstract}
Weakly-supervised semantic segmentation (WSSS) is introduced to narrow the gap for semantic segmentation performance from pixel-level supervision to image-level supervision.
Most advanced approaches are based on class activation maps (CAMs) to generate pseudo-labels to train the segmentation network.
The main limitation of WSSS is that the process of generating pseudo-labels from CAMs that use an image classifier is mainly focused on the most discriminative parts of the objects.
To address this issue, we propose Puzzle-CAM, a process that minimizes differences between the features from separate patches and the whole image.
Our method consists of a puzzle module and two regularization terms to discover the most integrated region in an object.
Puzzle-CAM can activate the overall region of an object using image-level supervision without requiring extra parameters.
In experiments, Puzzle-CAM outperformed previous state-of-the-art methods using the same labels for supervision on the PASCAL VOC 2012 dataset.
Code associated with our experiments is available at \url{https://github.com/OFRIN/PuzzleCAM}.

\end{abstract}
\begin{keywords}
Image segmentation,  Deep learning, Neural Networks, Weakly-supervised semantic segmentation
\end{keywords}
%

\section{Introduction}
\label{sec:intro}


Semantic segmentation is a fundamental approach using convolutional neural networks (CNNs) with the aim of correctly predicting the pixel-wise classification of an image.
Recently, fully-supervised semantic segmentation (FSSS) has achieved remarkable progress \cite{chen2018encoder, long2015fully, chen2014semantic}.
However, producing large-scale training datasets with precise pixel-level annotations per image is considerably expensive and requires labor-intensive and time-consuming tasks.
To solve this issue, many researchers have focused on weakly supervised semantic segmentation (WSSS), which is used to train networks using image-level annotations, scribbles, bounding boxes, and points.
Image-level supervision can be more easily conducted than other approaches in a group of weak supervision processes.
In this study, we only focused on learning semantic segmentation models using image-level supervision.


\begin{figure}[t]
\centering
\subfloat[]{\includegraphics[width=0.25\linewidth]{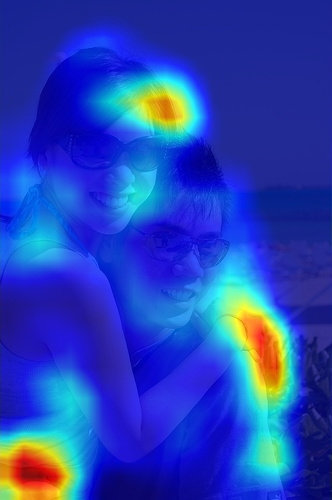}}  \hspace{2mm}
\subfloat[]{\includegraphics[width=0.25\linewidth]{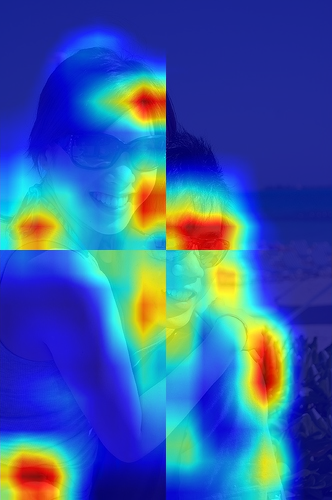}} \hspace{2mm}
\subfloat[]{\includegraphics[width=0.25\linewidth]{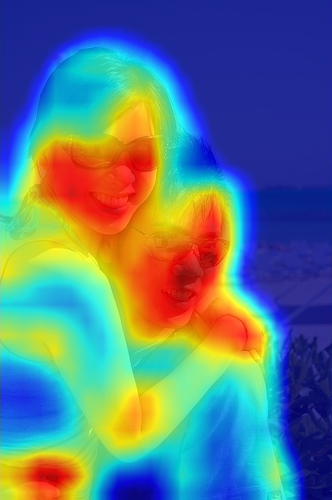}}
\caption{
CAMs generated from tiled and original images:
(a) conventional CAMs from the original image, (b) generated CAMs from the tiled images, and (c) predicted CAMs by the proposed Puzzle-CAM.
}
\label{fig:intro}
\vspace{-4mm}
\end{figure}

\begin{figure*}[t]%
\centering%
\includegraphics[width=\linewidth]{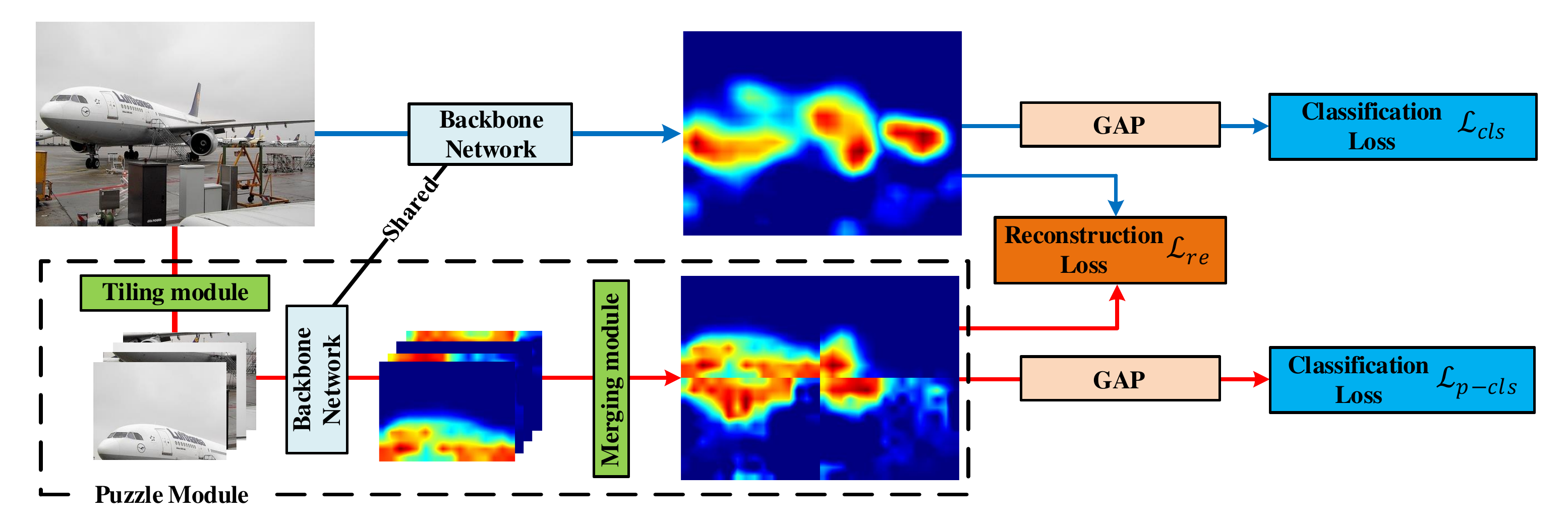}
\caption{
The overall architecture of the proposed Puzzle-CAM showing the integration of reconstructing regularization and the puzzle module.
}
\vspace{-4mm}
\label{fig:full}
\end{figure*}

Most previous methods \cite{ahn2018learning, Wang_2020_CVPR, lee2019ficklenet} using WSSS are based on the class activation maps (CAMs) \cite{zhou2016learning} to achieve good performance.
However, the generated CAMs are usually focused on small parts of the semantic objects to efficiently classify them, which prevents the segmentation models from learning pixel-level semantic knowledge.
Moreover, we can see that the CAMs generated from isolated patches in the tiled image are different from those gained from the original image.
As shown in Fig. \ref{fig:intro}, CAMs of the tiled image comprising tiled patches are significantly inconsistent compared to those of the original image.
The differences are factored in by enlarging the supervision gap between FSSS and WSSS by even more.


The above observations gave us the inspiration to address WSSS issues by using an attention-based feature learning method.
Thus, we propose Puzzle-CAM for WSSS training to detect integrated regions of objects.
Our method applies reconstructing regularization that corresponds to the generated CAMs from the tiled and original images to provide self-supervision.
To improve the network prediction consistency further, we introduced a puzzle module that splits the image and merges CAMs generated from the tiled image.
Puzzle-CAM consists of a Siamese neural network with reconstructing regularization loss that reduces the differences between the original and merged CAMs.
Our experiments yielded both quantitative and qualitative results that demonstrate the superiority of our approach. 


Our main contributions are as follows:
\begin{itemize}
  \item We propose Puzzle-CAM that incorporates reconstructing regularization with a puzzle module, to effectively enhance the quality of CAMs without adding layers.
  \item Puzzle-CAM outperformed existing state-of-the-art methods with the same level of supervision on the PASCAL VOC 2012 dataset. 
\end{itemize}

\section{Related Work}
\label{sec:related}


\subsection{Attention Mechanisms Using CNNs}
\label{ssec:attention}
These provides fine-grained information on the features learned in CNNs.
Simonyan \textit{et al.}~\cite{simonyan2013deep} used the error back-propagation strategy to visualize semantic regions whereas the combined attention model used the global average pooling (GAP) layer in the CNNs to generate the CAMs \cite{zhou2016learning} more efficiently.
Last, a final classifier is used to generate attention maps.
To the best of our knowledge, which attention mechanism is chosen does not have a great effect on achieving high performance with WSSS, and so we based Puzzle-CAM on the combined attention model because it is more manageable than the other attention mechanism.

\subsection{Weakly Supervised Semantic Segmentation}
\label{ssec:wsss}
Unlike FSSS, which requires pixel-wise labels for an image, WSSS employs lower level labeling, such as bounding boxes \cite{khoreva2017simple}, scribbles \cite{lin2016scribblesup}, and image-level classification labels \cite{ahn2018learning, lee2019ficklenet}.
Recently, the performance of WSSS has been significantly boosted by incorporating the CAMs.
Most previous WSSS methods refine the CAMs generated by the image classifier to approximate the segmentation mask \cite{ahn2018learning, ahn2019weakly, huang2018weakly, hou2018self, lee2019ficklenet}.
AffinityNet \cite{ahn2018learning} trains an additional network to learn similarities between the pixels, which often generates a transition matrix and multiplies with CAM to adjust its activation coverage. 
IRNet \cite{ahn2019weakly} generates a transition matrix from the boundary activation map and extends the method to achieve weakly supervised instance segmentation and WSSS. 
SEAM \cite{Wang_2020_CVPR} aims to refine CAMs using a pixel correlation module that captures context appearance information for each pixel and alters the original CAMs by using learned affinity attention maps.

\section{Methodology}
\label{sec:methodology}


\subsection{Motivation}
\label{ssec:motivation}

The conventional CAM of a single image highlights the most representative areas of each class. 
Therefore, when generating the CAM of the same class for image patches, the model focuses on finding the key features of the class using only the part of the object.
Thus, the merged CAM of image patches highlights the object area more accurately than the CAM of a single image. 
To take the aforementioned advantage, we propose Puzzle-CAM that trains a classifier using reconstruction loss to minimize the difference between the CAM of a single image and the merged CAM from image patches. 
By training a classification network using this reconstruction loss, the CAM covers more precisely the object area.
The Puzzle-CAM contains designed loss functions to match the CAMs generated from a tiled image with the original image (Fig. \ref{fig:full}).

\subsection{The Employed CAM Method}
\label{ssec:cam}
We first introduce the CAM method for producing the initial attention map.
Given feature extractor $F$ and classifier $\theta$, we can generate CAMs $A$ that are the collection of CAMs for all of the classes.
After training the classifier by image-level supervision, we apply the weights of the $c$-channel classifier as $\theta^{c}$ on feature map $f = F(I)$ from input image $I$ to obtain the CAM of class $c$ as follows: 

\begin{equation} \label{eq1}
A_c = \theta_{c}^{\top}f.
\end{equation}

The generated CAM is normalized by using the maximum value of $A_c$.
Finally, we obtain the CAMs for all of the classes ($A$) by concatenating $A_c$. 

\subsection{The Puzzle Module}
\label{ssec:module}
When matching partial and full features, the key is to narrow the gap between FSSS and WSSS.
The puzzle module consists of tiling and merging modules.
From input image $I$ of size $W\times H$, the tiling module generates non-overlapping a tiled patches $\{I^{1,1}, I^{1,2}, I^{2,1}, I^{2,2}\}$ of size $W/2 \times H/2$.
Next, we generate $A^{i,j}$ CAMs for each $I^{i,j}$.
Finally, the merging module attaches all $A^{i,j}$ into a single CAMs $A^{re}$ that has the same shape as $A^s$ which is the CAMs of $I$.

\subsection{The Loss Design for Puzzle-CAM}
\label{ssec:loss}

We employed a GAP layer at the end of the network to incorporate prediction vector $\hat{Y} = \sigma(G(A_c))$ for image classification and to adopt multi-label soft margin loss for the classification task.
For notational convenience, we define $Y_t$ as

\begin{equation}\label{eq2}
    \hat{Y}_t = 
\begin{cases}
    \hat{Y}      ,& \text{if } Y = 1 \\
    1 - \hat{Y},  & \text{otherwise}
\end{cases}
\end{equation}

\begin{equation}\label{eq3}
\ell_{cls}(\hat{Y}, Y) = -log(Y_t).
\end{equation}


The CAMs of the original ($A^s$) and tiled ($A^{re}$) images are converted using the GAP layer with prediction vectors $\hat{Y}^{s} = G(A^{s})$ and $\hat{Y}^{re} = G(A^{re})$, respectively. 
The classification losses for the original and reconstructed images are respectively calculated as

\begin{align}\label{eq4}
\mathcal{L}_{cls} &= \ell_{cls}(\hat{Y}^{s}, Y), \\  
\mathcal{L}_{p-cls} &= \ell_{cls}(\hat{Y}^{re}, Y).
\end{align}

These two classification losses are used to improve the performance of the image classification. 
To reinforce the CAMs from the original image, we added reconstructing regularization to correspond with the original and reconstructed CAMs. 
The reconstruction loss for the original CAM can be easily defined as

\begin{equation} \label{eq6}
\mathcal{L}_{re} = \left\|{A^{s} - A^{re}} \right\|_{1}.
\end{equation}

In summary, the final loss can be written as 

\begin{equation} \label{eq7}
\mathcal{L} = \mathcal{L}_{cls} + \mathcal{L}_{p-cls} + \alpha \mathcal{L}_{re}.
\end{equation}

where $\alpha$ is the balance of the weights for the different losses.
The classification losses, $\mathcal{L}_{cls}$ and $\mathcal{L}_{p-cls}$, are used to roughly estimate the region of the object. 
The reconstruction loss, $\mathcal{L}_{re}$, is used to narrow the gaps between the pixel- and image-level supervision processes. 
We report details of the network training settings and probe into the effectiveness of the proposed module in the experiments section.

\section{Experimental Results}
\label{sec:experiments}
\begin{table}[t]
\caption{
Ablation study of the Puzzle-CAM loss functions using ResNet-50 as the backbone.
}
\footnotesize
\centering
{
\begin{tabu} to \linewidth{X[c,0.8] X[c,0.8] X[c,0.8] | X[c,1.2] } \hline \hline
$L_{cls}$  & $L_{p-cls}$      & $L_{re}$ & mIoU (\%)        \\ \hline
\checkmark & & & 47.82  \\
\checkmark &\checkmark & & 47.70   \\ 
\checkmark & & \checkmark & 49.21  \\  
\checkmark & \checkmark & \checkmark & \textbf{51.53}   \\ \hline\hline

\end{tabu}
}
\label{tb:ablation_for_loss}
\vspace{-5mm}
\end{table}

\subsection{Implementation Details}
\label{ssec:implementation}

We evaluated our method using the PASCAL VOC 2012 dataset \cite{everingham2010pascal}.
The dataset was separated into 1,464 images for training, 1,449 for validation, and 1,456 for testing. 
Following the experimental protocol used in previous methods \cite{ahn2018learning, Wang_2020_CVPR, lee2019ficklenet}, we took additional annotations from the Semantic Boundary Dataset \cite{hariharan2011semantic} to build an augmented training set with 10,582 images. 
The images were randomly re-scaled in the range of [320, 640] and then cropped to $512 \times 512$ as the network inputs. 
For all experiments, we set $\alpha = 4$ as the maximum and linearly ramped up $\alpha$ to its maximum value by half epochs.
During inference, we utilized classifiers without the puzzle module.
Thus, we adopted multi-scale and horizontal flip to generate pseudo-segmentation labels.
We used four TITAN-RTX GPUs for the model to train the dataset.



\subsection{Ablation Studies}
\label{ssec:ablation}

We conducted ablation studies on the main components of Puzzle-CAM by applying the mean Intersection-Over-Union (mIoU) metric (Table~\ref{tb:ablation_for_loss}), for which the baseline was $mIoU = 47.82\%$. 
With the proposed reconstructing regularization ($\mathcal{L}_{re}$) of the  tiled patches, the baseline was boosted to $mIoU = 49.21\%$, while the proposed classification loss from the tiled patches ($\mathcal{L}_{p-cls}$) was similar to the baseline. 
Both $\mathcal{L}_{re}$ and $\mathcal{L}_{p-cls}$ consistently improved the baseline by a $3.71\%$. 

\begin{figure}[t]
\centering
\subfloat[$L_{cls}$]{\includegraphics[width=0.45\linewidth,height=2.5cm]{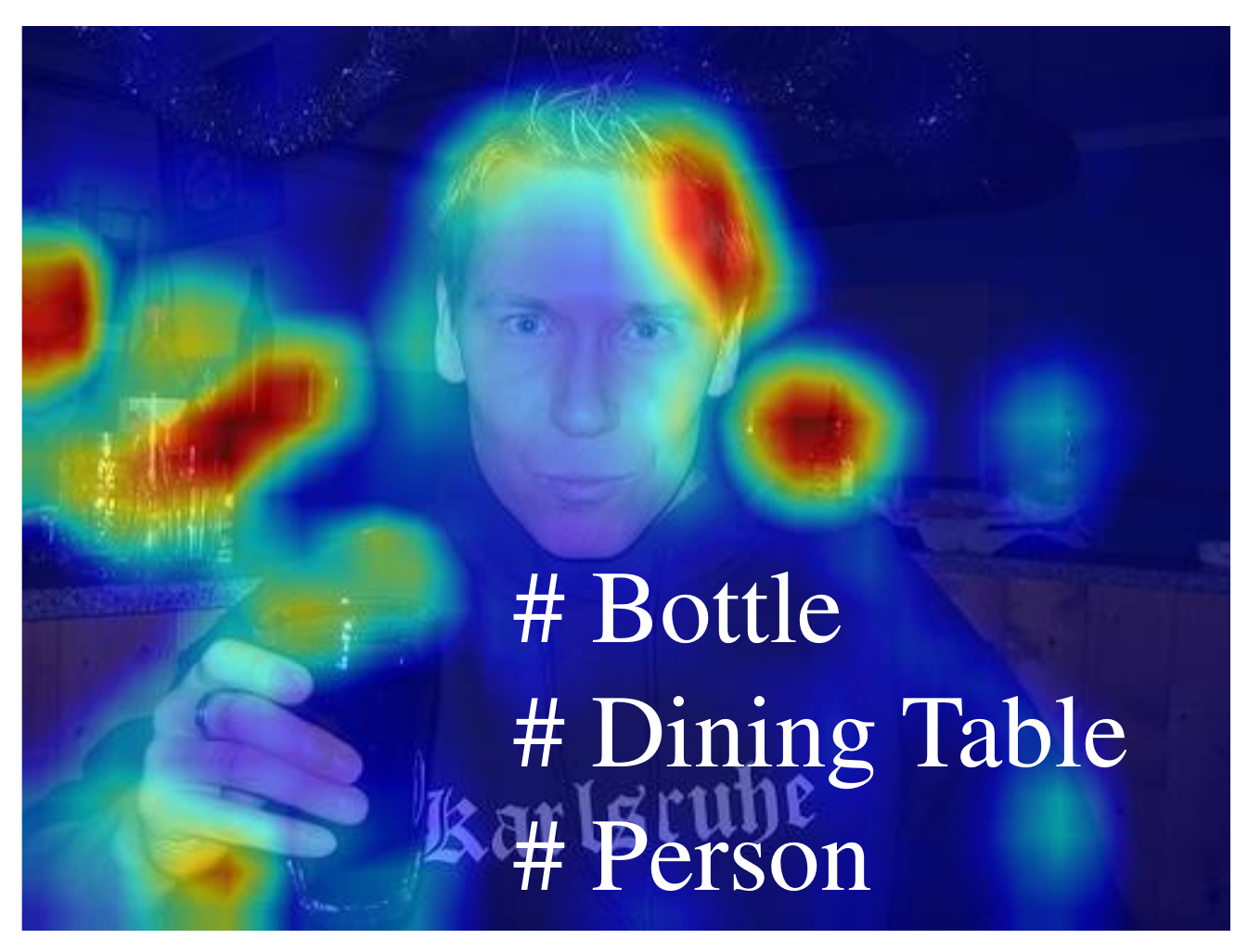}} \hspace{2mm}
\subfloat[$L_{cls} + L_{p-cls}$]{\includegraphics[width=0.45\linewidth,height=2.5cm]{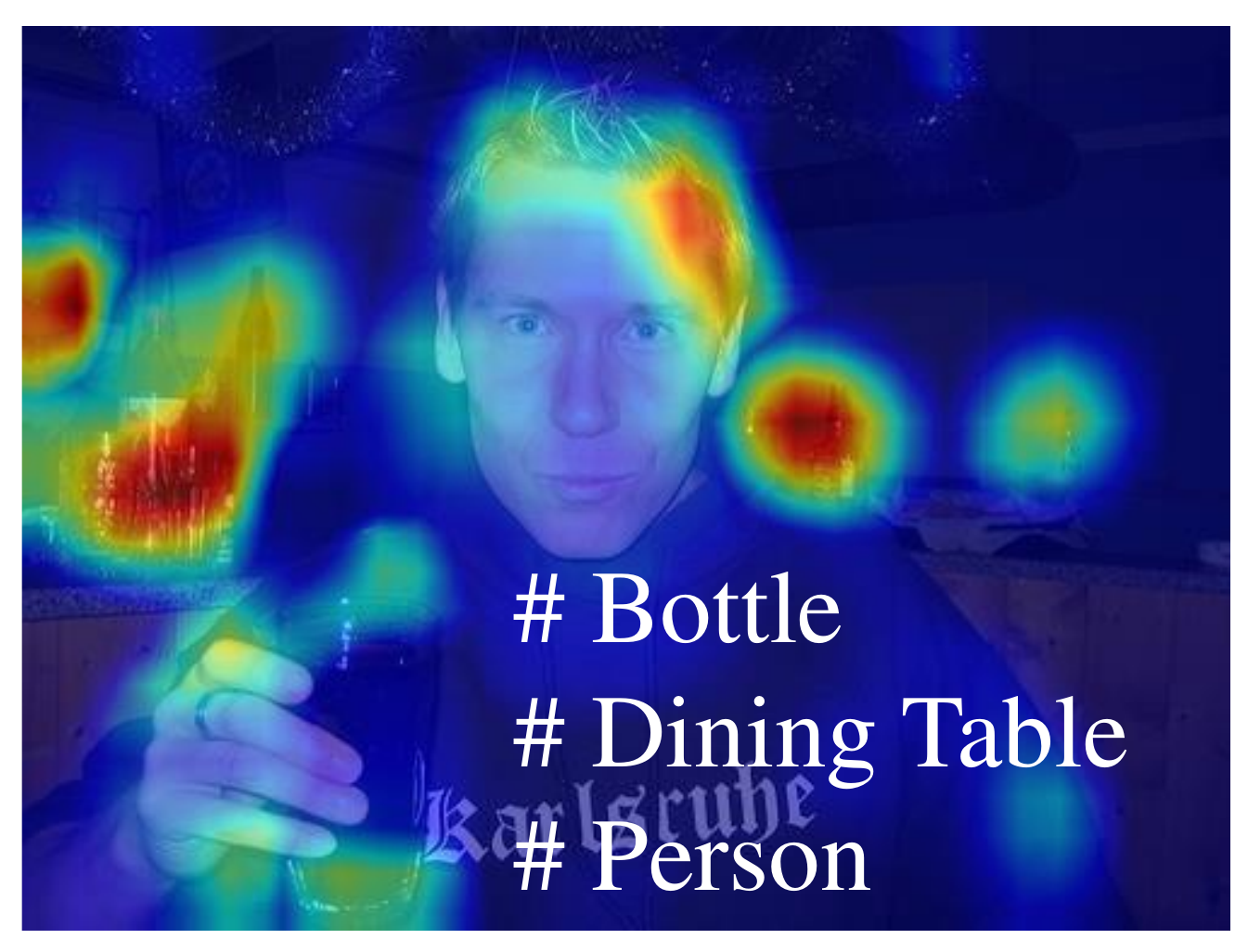}}\\ 
\subfloat[$L_{cls} + L_{re}$]{\includegraphics[width=0.45\linewidth,height=2.5cm]{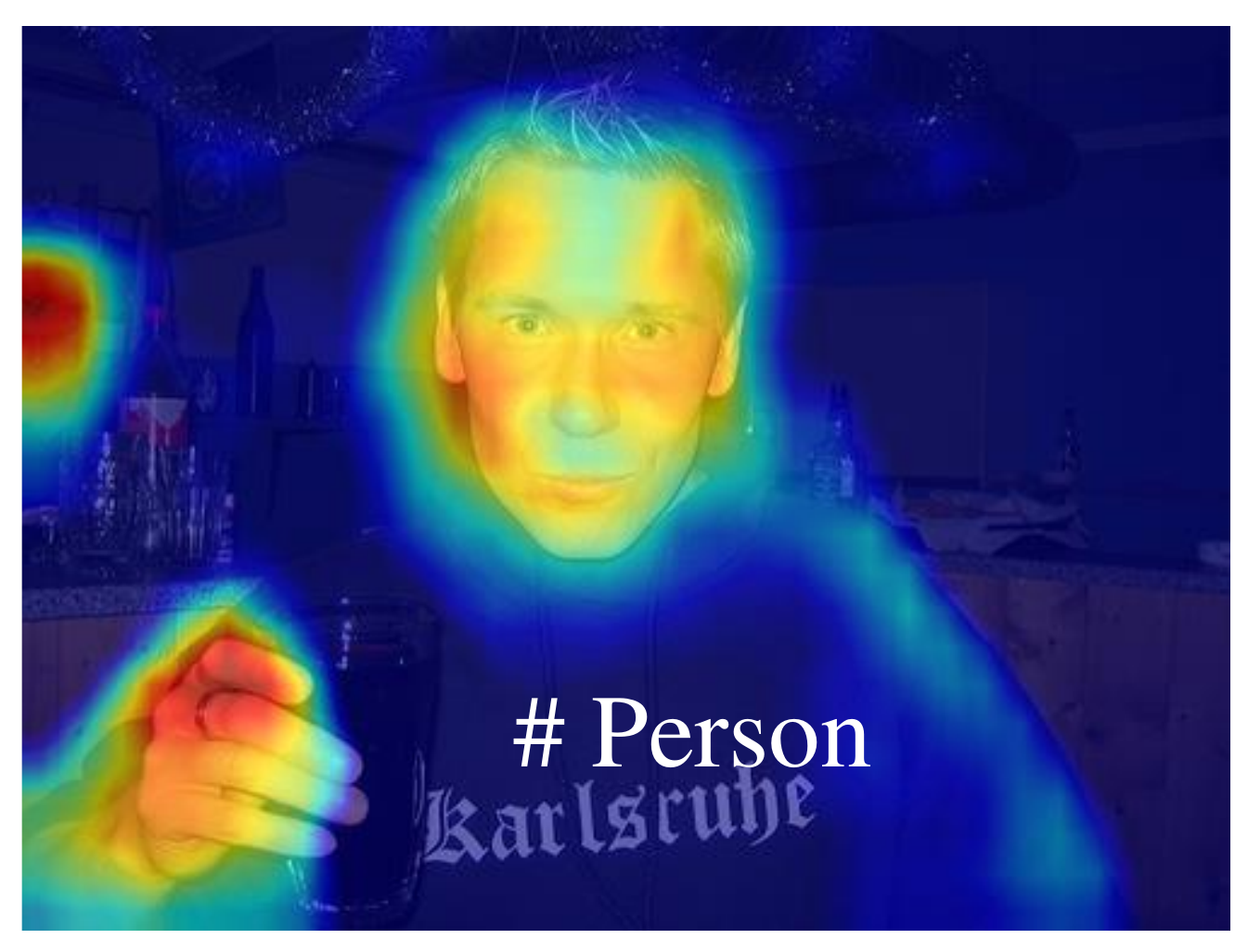}} \hspace{2mm}
\subfloat[$L_{cls} + L_{p-cls} + L_{re}$]{\includegraphics[width=0.45\linewidth,height=2.5cm]{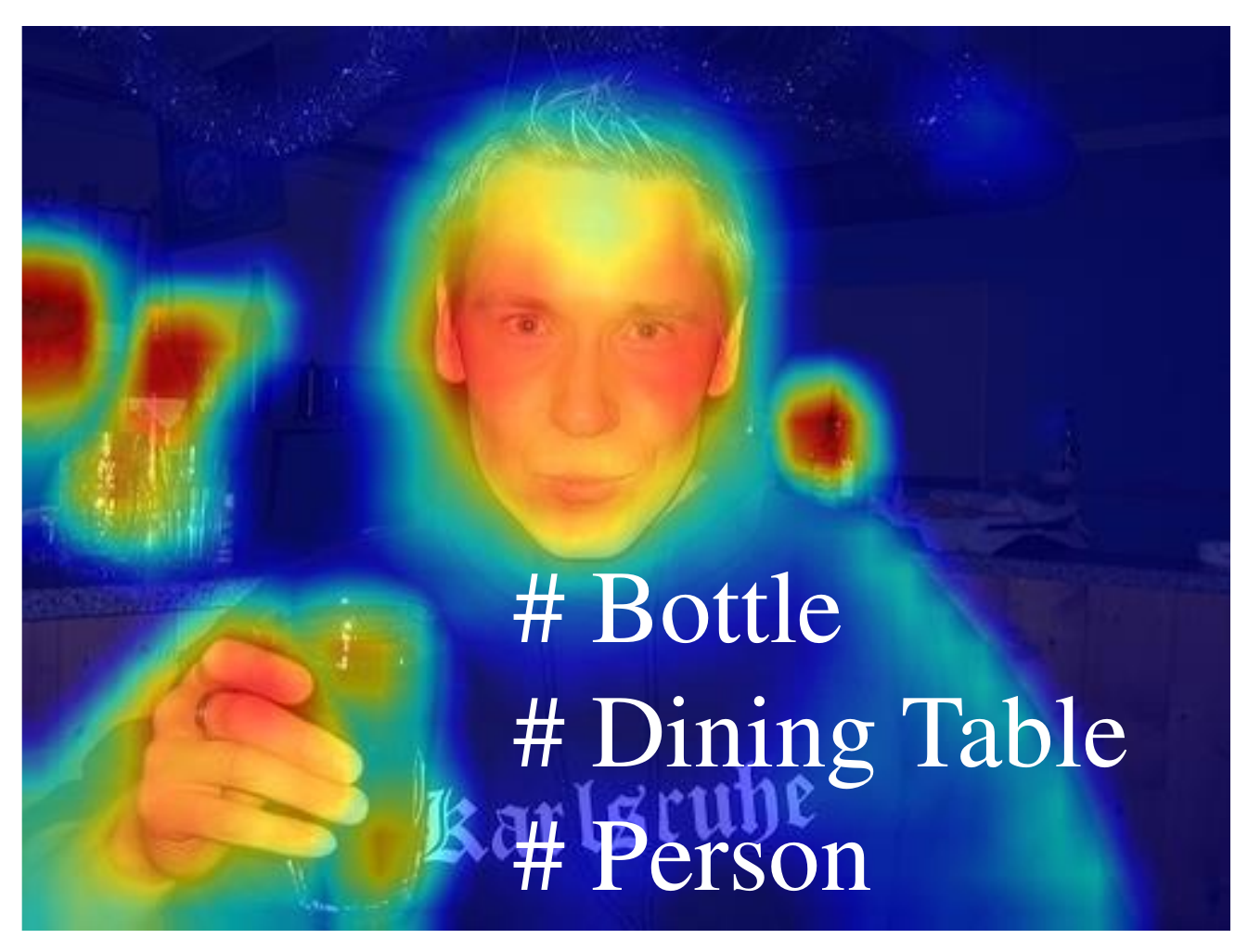}}
\caption{
    Visualization of the prediction tags and CAMs by using combinations of loss functions. 
    In (d), the final CAMs not only suppressed over-activation but also expanded the CAMs into complete object activation coverage . 
}
\label{fig:ablation}
\vspace{-2mm}
\end{figure}

We visualized the CAMs by using combinations of loss functions for each of them (see Eq. \ref{fig:ablation}).
When the classification losses ($L_{p-cls}$) only are employed, the result will show no marginal differences.
Meanwhile, when the reconstruction loss ($L_{re}$) only is employed, the result will show improved localization ability for some classes compared to the original but the method will fail to predict several classes.
When both sets of losses are combined, the result will show improved localization without suffering classification losses.




\begin{table}[t]
\caption{
Quality of the pseudo semantic segmentation labels in mIoU, evaluated on the PASCAL VOC 2012 training set \cite{everingham2010pascal}. 
RW, random walk with AffinityNet \cite{ahn2018learning}; dCRF, dense conditional random field \cite{krahenbuhl2011efficient}.
}
\footnotesize
\centering
{
\begin{tabu} to \linewidth{X[c,1.6] | X[c,1.5] | X[c,0.5] | X[c,1.0] | X[c,1.3] }  \hline \hline
\multirow{2}{*}{Method} & \multirow{2}{*}{Backbone} & CAM & CAM & CAM+RW \\ 
& & (\%) & +RW (\%) & +dCRF (\%) \\ \hline
AffinityNet \cite{ahn2018learning} & ResNet-50 & 47.82 & 58.10 & 59.70 \\
Puzzle-CAM & ResNet-50 & 51.53 & 64.16  & 64.70 \\ 
Puzzle-CAM & ResNeSt-50 & 57.59 & 69.48  & 69.91 \\ 
Puzzle-CAM & ResNeSt-101 & 61.85 & 71.92 & 72.46 \\ 
Puzzle-CAM & ResNeSt-269 & \textbf{62.45} & \textbf{74.14} & \textbf{74.67} \\ \hline\hline
\end{tabu}
}
\vspace{-4mm}
\label{tb:compare_cam_with_rw}
\end{table}

\begin{table}[t]
\caption{
Comparison of Puzzle-CAM and existing state-of-the-art methods on the PASCAL VOC 2012 $val$ and $test$ datasets. 
$\mathcal{I}$, image-level labels;  $\mathcal{S}$, external saliency models.
}
\footnotesize
\centering
{
\begin{tabu} to \linewidth{X[c,1.2] X[c,1.0] X[c,0.75]  | X[c,0.2]  X[c,0.2] }  \hline \hline
Method  & Backbone & Supervision & val & test \\ \hline
AffinityNet~\cite{ahn2018learning} & Wide-ResNet-38 & $\mathcal{I}$ & 61.7 & 63.7 \\
DSRG~\cite{huang2018weakly} & ResNet-101 & $\mathcal{I}$ + $\mathcal{S}$ & 61.4 & 63.2 \\ 
SeeNet~\cite{hou2018self} & ResNet-101 & $\mathcal{I}$ + $\mathcal{S}$ & 63.1 & 62.8 \\
IRNet~\cite{ahn2018learning} & ResNet-50 & $\mathcal{I}$ & 63.5 & 64.8 \\
FickleNet~\cite{lee2019ficklenet} & ResNet-101 & $\mathcal{I}$ + $\mathcal{S}$ & 64.9 & 65.3 \\ 
ICD~\cite{fan2020learning} & ResNet-101 & $\mathcal{I}$ & 64.1 & 64.3 \\ 
SEAM~\cite{Wang_2020_CVPR} & Wide-ResNet-38 & $\mathcal{I}$ & 64.5 & 65.7 \\ \hline
Ours (Puzzle-CAM) & ResNeSt-101 & $\mathcal{I}$ & 66.9 & 67.7 \\
Ours (Puzzle-CAM) & ResNeSt-269 & $\mathcal{I}$ & \textbf{71.9} & \textbf{72.2} \\

 \hline\hline
\end{tabu}
}
\label{tb:final}
\end{table}

\begin{figure}
\begin{minipage}[b]{1.0\linewidth}
  \centering
  \centerline{\includegraphics[width=1.0\linewidth]{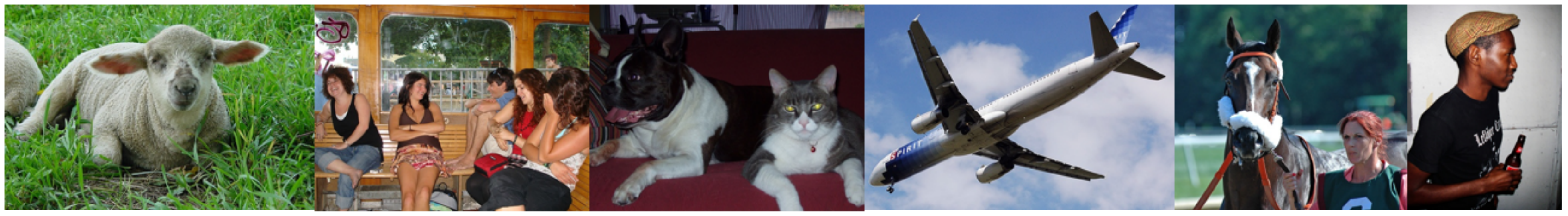}}
\end{minipage}

\begin{minipage}[b]{1.0\linewidth}
  \centering
  \centerline{\includegraphics[width=1.0\linewidth]{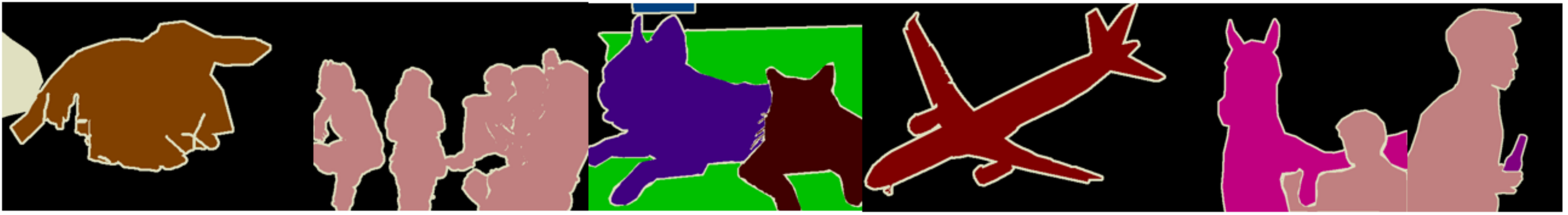}}
\end{minipage}

\begin{minipage}[b]{1.0\linewidth}
  \centering
  \centerline{\includegraphics[width=1.0\linewidth]{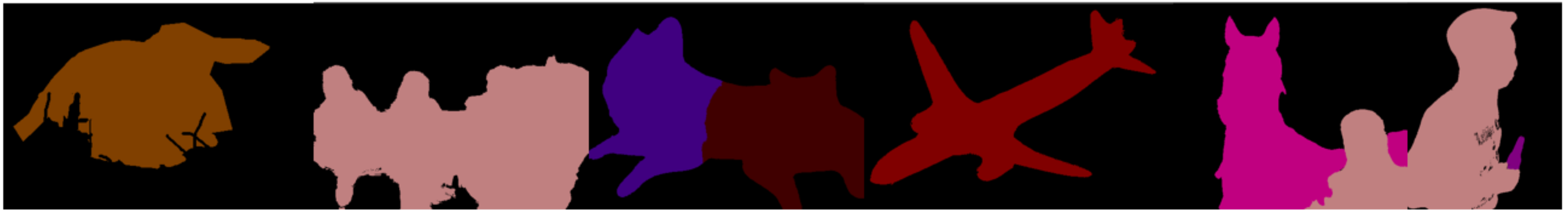}}
\end{minipage}

\caption{Qualitative segmentation results on the PASCAL VOC 2012 \textit{val} set. Top: original images. Middle: ground truth. Bottom: prediction of the segmentation model trained using the pseudo-labels from Puzzle-CAM.}
\label{fig:result}
\vspace{-4mm}
\end{figure}

\subsection{Comparison with Existing State-of-the-Art Methods}
\label{ssec:comparisons}

To further improve the accuracy of pseudo pixel-level annotations, we followed the approach in \cite{ahn2018learning} to train AffinityNet based on Puzzle-CAM. 
We adopted ResNeSt architecture that universally improves the learned feature representations to boost performance across image classification, object detection, instance segmentation and semantic segmentation. 
In Table~\ref{tb:compare_cam_with_rw}, we report the performances with the original CAMs used by the baseline AffinityNet \cite{ahn2018learning} and Puzzle-CAM.

The final synthesized pseudo-labels achieved $74.67\%$ mIoU on the PASCAL VOC 2012 $train$ set.
Puzzle-CAM was then used to train the segmentation model DeepLabv3+ \cite{chen2018encoder} with the ResNeSt-269 \cite{zhang2020resnest} backbone using the pseudo-labels in full supervision mode to achieve the final segmentation results. 
Table \ref{tb:final} reports a comparison of the mIoU values for the proposed method and the previous approaches. 
Compared to the baseline methods, Puzzle-CAM had remarkably improved performances on both the $val$ and $test$ sets with the same settings for training.
Fig. \ref{fig:result} shows some qualitative results on the $val$ set that illustrate that the proposed method worked well on both large and small objects.




\section{Conclusions}
\label{sec:conclusion}

In this paper, we proposed the Puzzle-CAM algorithm to narrow the supervision gap between FSSS and WSSS using image-level labels. 
To improve the network for generating consistent CAMs, we designed a puzzle module and adopted reconstructing regularization to match partial and full features.
Not only did Puzzle-CAM consistently generate features from local tiled patches but it also fitted the shape of the ground truth masks better. 
The segmentation network trained by our synthesized pixel-level pseudo-labels achieved state-of-the-art performance on the PASCAL VOC 2012 dataset, which proves the effectiveness of our approach. 
We believe that the concepts of Puzzle-CAM as a training module can be generalized and will benefit other weakly- and semi-supervised tasks, such as semantic and instance segmentation.

\textbf{Acknowledgement} 
This work was supported in part by the National Research Foundation of Korea (NRF) under Grant NRF-2020M3C1C2A01080819, and in part by the Ministry of Interior and Safety (MOIS), Korea Government under Grant 2019-MOIS33-004.  




\bibliographystyle{IEEEbib}

\end{document}